\documentclass[10pt,twocolumn,letterpaper]{article}

\usepackage[pagenumbers]{cvpr} 

\usepackage{graphicx}
\usepackage{amsmath}
\usepackage{amssymb}
\usepackage{booktabs}
\usepackage{enumitem}
\usepackage[font=small,skip=0pt]{caption}
\usepackage{algorithm,algorithmic}

\usepackage[pagebackref,breaklinks,colorlinks]{hyperref}

\newcommand{\RR}{\mathbb{R}}

\newcommand{\quantize}{\mathbf{q}}

\newcommand{\Lsize}{1024$\times$512}
\newcommand{\Msize}{512$\times$256}
\newcommand{\Ssize}{256$\times$128}
\newcommand{\vqganOne}{VQGAN\text{$_{1}$}}
\newcommand{\vqganTwo}{VQGAN\text{$_{2}$}}

\usepackage[capitalize]{cleveref}
\crefname{section}{Sec.}{Secs.}
\Crefname{section}{Section}{Sections}
\Crefname{table}{Table}{Tables}
\crefname{table}{Tab.}{Tabs.}

\def\cvprPaperID{0000} 
\def\confName{CVPR}
\def\confYear{2022}

\begin{document}

\title{Diverse Plausible 360-Degree Image Outpainting\\
for Efficient 3DCG Background Creation}

\author{Naofumi Akimoto \qquad Yuhi Matsuo \qquad Yoshimitsu Aoki\\
Keio University
}

\twocolumn[{%
\renewcommand\twocolumn[1][]{#1}%
\maketitle 
\vspace{-10mm}
\begin{center}
    
    \includegraphics[width=\textwidth]{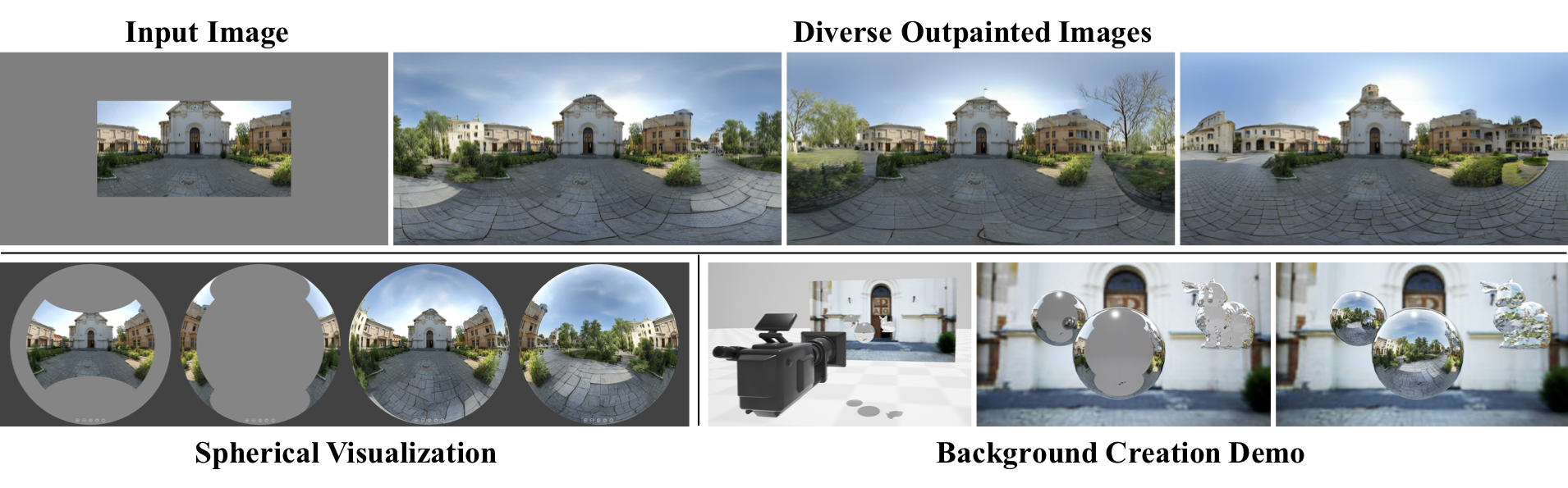}
    \captionof{figure}{We generate the plausible environment from a narrow field of view image, using a transformer-based outpainting method that considers the nature of 360-degree images, to realize efficient 3DCG scene creation. See also demonstrations in the supplementary video. }
    \vspace{2mm}
\label{fig:teaser}
\end{center}%
}]


\begin{abstract}
   We address the problem of generating a 360-degree image from a single image with a narrow field of view by estimating its surroundings. Previous methods suffered from overfitting to the training resolution and deterministic generation. This paper proposes a completion method using a transformer for scene modeling and novel methods to improve the properties of a 360-degree image on the output image. Specifically, we use CompletionNets with a transformer to perform diverse completions and AdjustmentNet to match color, stitching, and resolution with an input image, enabling inference at any resolution. To improve the properties of a 360-degree image on an output image, we also propose WS-perceptual loss and circular inference. Thorough experiments show that our method outperforms state-of-the-art (SOTA) methods both qualitatively and quantitatively. For example, compared to SOTA methods, our method completes images 16 times larger in resolution and achieves 1.7 times lower Fréchet inception distance (FID). Furthermore, we propose a pipeline that uses the completion results for lighting and background of 3DCG scenes. Our plausible background completion enables perceptually natural results in the application of inserting virtual objects with specular surfaces.
\end{abstract}

\section{Introduction}
\label{sec:introduction}

In recent three-dimensional computer graphics (3DCG) production, 360-degree images are helpful for efficiently creating lighting and backgrounds. For example, a designer might spend much time creating 3D objects in the near field and creating the background quickly by using 2D images with a narrow field of view (NFoV) images or 360-degree images. However, the production method of creating the background by placing 2D images behind a 3D object cannot fully represent the scenery reflected on the surface of the 3D object. Of course, this problem does not occur if the image surrounds the object in 360 degrees. However, 360-degree images, especially high dynamic range images (HDRI), are generally more expensive to prepare than NFoV images.

This paper addresses the problem of converting an NFoV image into a 360-degree image by complementing its surroundings to obtain a 360-degree environment consistent with the image given as a partial background (Fig.~\ref{fig:teaser}). By solving this problem, users can use only a NFoV image to reflect the surrounding environment to objects \cite{akimoto2019360, somanath2021hdr}, or in the case of HDRI, to achieve natural shadows and global illumination through Image-Based Lighting \cite{somanath2021hdr, gardner-sigasia-17}.

\begin{figure}[t]
  \centering
   \vspace{-1mm}
   \includegraphics[width=\linewidth]{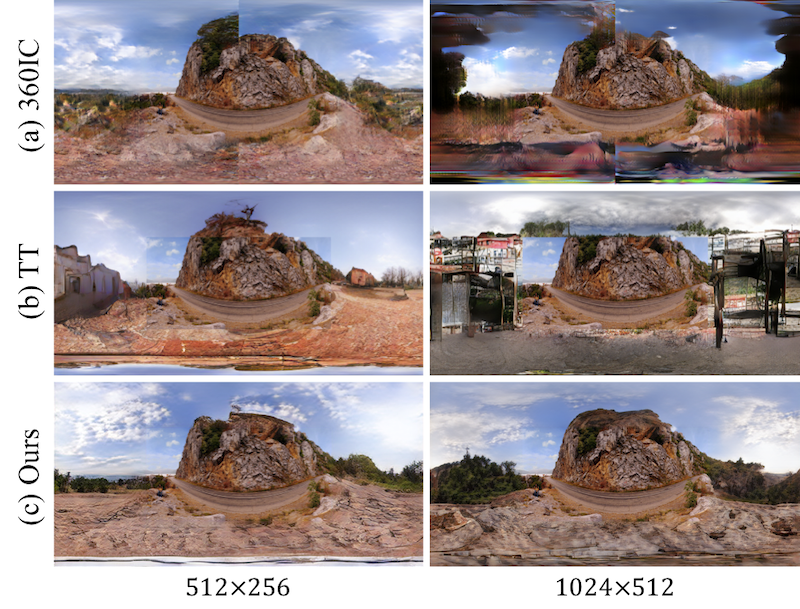}
   \caption{Limitations of prior methods. (a) CNN-based method \cite{akimoto2019360} and (b) transformer-based translation method \cite{Esser_2021_CVPR} suffer from overfitting to resolution during training (\Msize{}). Furthermore, (b) has no connection between the ends.}
   \label{fig:limitations_prior}
\end{figure}

For use by designers, it is desirable to infer NFoV images of any size and to have choices by generating diverse 360-degree images. However, existing methods are deterministic in their estimation and can only correctly infer the trained resolution. For example, as shown in Fig.~\ref{fig:limitations_prior}, 360IC\cite{akimoto2019360}, trained at \Msize{}, produces many artifacts at \Lsize{}. We believe that this is due to distortions caused by the equirectangular projection (ERP).
Based on the above, our goal is to realize the outpainting of 360-degree images with the following two properties. (1) Sample diverse outputs for a single input and (2) infer arbitrary resolutions.

The key idea of our approach is to introduce a transformer \cite{vaswani2017attention} into an outpainting method for diverse outputs. In previous works, TT \cite{Esser_2021_CVPR} is an image-to-image translation method with a transformer that can generate various outputs through sampling from the learned distribution. However, as shown in Fig.~\ref{fig:limitations_prior}(b), TT alone is not consistent enough with the input image, and it can only generate images of fixed size. Therefore, we cannot directly introduce TT into our task, and the resolution problem remains.
To solve this technical problem of introducing a transformer, we propose an additional network as a second stage. Specifically, we present a framework comprising CompletionNets and AdjustmentNet.
(1) CompletionNets is an image completion module that uses the same networks as TT and two novel techniques to be proposed later. (2) AdjustmentNet improves the consistency of color, stitching, and resolution between the output result of CompletionNets and the input image. Because AdjustmentNet adjusts the fixed-size output of CompletionNets to the size of the input image, we can obtain completion results for any image resolution.

Furthermore, because the above framework does not yet sufficiently consider the unique properties of 360-degree images, we propose two novel techniques for this purpose.
First, to achieve continuity at both ends of an image, which is a property of 360-degree images, we propose circular inference as a new auto-regressive order for a transformer. It improves the connectivity at both ends of an image at the pixel and semantic levels by performing inference while circulating the image. Second, to further improve the perceptual quality, we propose a WS-perceptual loss function for training of CompletionNets. This loss function reflects that 360-degree images have different information content along the latitudinal direction and improves the performance of 360-degree image modeling by focusing on computing the loss in the information-rich regions.

Our thorough experiments show not only that the proposed method can perform diverse completions at arbitrary resolutions but also that the proposed method outperforms several state-of-the-art methods both qualitatively and quantitatively. For example, in terms of FID score, our method shows 1.7\% lower improvement than 360IC and achieves plausible completion for images with 16 times as many pixels (\Lsize) as EnvMapNet \cite{somanath2021hdr} (\Ssize).

Moreover, we propose a pipeline to create an HDR environment map from a single NFoV image and use it as lighting and background in 3DCG. Through demonstrations, we show that our method reaches the quality of 360-degree image completion, which can be used for 3DCG and is helpful for efficient background creation.

The proposed method produces a plausible 360-degree image and provides various completion results, allowing designers to choose their preferred result among them. Considering these characteristics, we conclude with a discussion of potential applications.

Our contributions can be summarized as follows: 
\begin{itemize}[noitemsep,topsep=0pt]
\itemsep0em 
\item We propose AdjustmentNet to introduce a transformer into 360-degree image outpainting, which enables diverse and arbitrary-resolution outputs.
\item We propose two novel techniques for acquiring the properties of 360-degree images: WS-perceptual loss for the training of CompletionNets and circular inference for the transformer. These allow us to outperform previous methods both quantitatively and qualitatively.
\item We demonstrate that our high-resolution and plausible completion renders natural-looking scenes even when specular virtual objects are close to a camera or the camera views all around on 3DCG scenes.
\end{itemize}

\begin{figure*}[t!]
  \centering
  \vspace{-3mm}
  \begin{minipage}{0.64\linewidth}
    \includegraphics[width=0.99\linewidth]{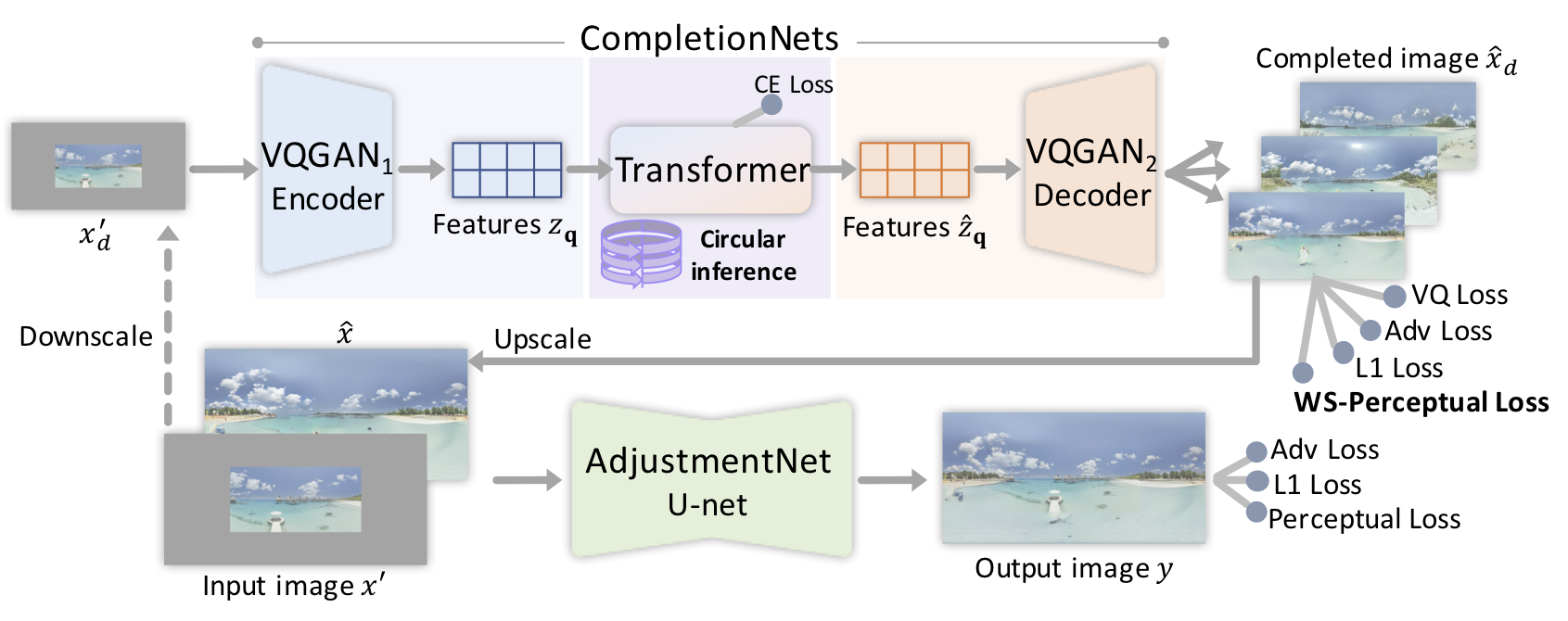} 
  \end{minipage}
  \begin{minipage}{0.015\linewidth}
  \end{minipage}
  \begin{minipage}{0.34\linewidth}
    \caption{\small\textbf{Framework overview.} Our method comprises two modules: CompletionNets and AdjustmentNet. CompletionNets can sample various completion results from a fixed-size image input. AdjustmentNet improves the consistency of color, stitching, and resolution between the CompletionNets' output and the input image, resulting in completion for any image size.
    }
    \label{fig:framework} 
  \end{minipage}
\vspace{-5mm}
\end{figure*}

\section{Related Work}
\label{sec:related}
{\bf Image completion}. 
Image inpainting is the task of filling in missing regions with appropriate pixels \cite{barnes2009patchmatch,Hays2007scene}. Learning-based image inpainting \cite{pathak2016context,IizukaSIGGRAPH2017,li2017generative}, which is CNN-trained on large datasets, has been extensively studied recently. In addition, attention-based image inpainting has been proposed and shown promising results \cite{yu2018generative,liu2019coherent}. However, most of the methods that train CNN with GAN produce deterministic outputs. In other words, these methods output only one result for an input image. PIC \cite{zheng2019pluralistic}, in contrast, outputs multiple results by employing CVAE \cite{sohn2015learning}.

Image outpainting is an extrapolation problem that generates the surroundings of an input image. This task can include image extension \cite{wang2019wide,teterwak2019boundless}, novel view synthesis \cite{wiles2020synsin}, infinite landscape generation \cite{yang2019very,liu2021infinite}, and panorama generation \cite{lin2019coco,Skorokhodov_2021_ICCV}. A similar task to our work is the generation of panoramas that provide a 360-degree view. However, ERP images have significant distortions at the top and bottom of the images and, therefore, panoramic images cannot be used as ERP images.

Using a transformer for image completion is another recently studied topic \cite{chen2020generative, Esser_2021_CVPR, Wan_2021_ICCV} and has the following two main advantages. (1) Non-local attention can help generate a global structure and contextual consistency. (2) Sampling from the distribution learned by a transformer leads to more diverse completions than CVAE, as verified by \cite{Wan_2021_ICCV}. In contrast, the disadvantages of a transformer are that it requires an enormous computational cost when dealing with large images. Moreover, transformer-based image-to-image translation (Im2Im) \cite{Esser_2021_CVPR, rombach2021geometryfree} resamples the pixels in the conditional input region, resulting in a loss of consistency with the original pixels.

However, the above image completion works designed their approach for planar images. 
In other words, the generated results lacked the properties of 360-degree images, such as the connection between the two ends of the images and the latitudinal distortion caused by the projection. In contrast, our method can plausibly complete a 360-degree image by introducing a transformer into an image outpainting while considering the 360-degree properties.

{\bf 360-Degree image outpainting}.
360-Degree image outpainting is the task of completing the surroundings of a partial 360-degree image. Inverse rendering \cite{neuralSengupta19,wang2021learning} and lighting estimation \cite{gardner-sigasia-17, Gardner_2019_ICCV,song2019neural,legendre2019deeplight} perform the task of completing a 360-degree image as an intermediate task to represent lighting through the 360-degree image (environment map). However, these methods cannot predict high-frequency textures, and the image resolution is small. Similar to our work is the task focused on pixel completion \cite{akimoto2019360, hara2021spherical, somanath2021hdr}. These are image completion methods that consider the properties of 360-degree images. 360IC and SIG-SS \cite{hara2021spherical} employ techniques to improve the continuity of both ends of a 360-degree image. EnvMapNet trains a network by weighting the pixel loss to account for the difference in latitudinal information density due to the projection. However, except for SIG-SS, which uses CVAE to sample the strength of symmetry, these methods are deterministic outputs. Furthermore, they suffer from overfitting for image resolution during training. 


\section{Method}
\label{sec:method}
We generate a 360-degree image by completing the surrounding area of an NFoV image. In this work, 360-degree images are ERP images. Again, our goal is to obtain multiple and diverse outputs for a single input image and enable inference at arbitrary resolutions different from the training resolution. 
Our approach is to perform diverse completions of a scene using a transformer. TT has already shown that diverse completion is possible with Im2Im using a transformer. However, as mentioned in Sec.~\ref{sec:introduction} and Fig.~\ref{fig:limitations_prior}, the transformer-based Im2Im is not suitable for 360-degree images in the terms of overfitting a training resolution and consistencies with the input pixels. Therefore, we propose a framework extended with AdjustmentNet to solve the problems (Sec.~\ref{sec:model}). Moreover, we propose a new loss, WS-perceptual loss (Sec.~\ref{sec:training}), and a new inference method for the transformer, circular inference (Sec.~\ref{sec:inference}), to reflect the properties of 360-degree images to outputs.

{\bf Overview.} Fig.~\ref{fig:framework} shows an overview of our proposed framework. The input of the entire framework is an incomplete image $x' \in \RR^{H \times W \times 3}$. During training, we crop some regions from the ERP images $x \in \RR^{H \times W \times 3}$ and fill the remaining regions with gray values. The output $y \in \RR^{H \times W \times 3}$ is a restored image of the complete 360-degree scene.

Our method comprises two modules: CompletionNets and AdjustmentNet.
First, we downsize the incomplete input image $x'$ to a fixed size and use it as input to CompletionNets. CompletionNets completes the incomplete image $x'_d  \in \RR^{h \times w \times 3}$ using a transformer. Because the completed image $\hat{x}_d \in \RR^{h \times w \times 3}$ is of fixed size, we restore it by upscaling to the original size of the input image. Next, to enable inference at arbitrary resolutions different from the training resolution, AdjustmentNet uses the completed image $\hat{x} \in \RR^{H \times W \times 3}$ and the input image $x'$ to estimate the high-frequency texture of the completed image $y$ according to the input image $x'$. It also performs stitching and color correction to obtain the final output.

\subsection{Model Architecture} \label{sec:model}
{\bf CompletionNets.} 
The primary network structure of CompletionNets is the same as TT: two VQGANs \cite{Esser_2021_CVPR} and a transformer. The approach of TT is vector-quantized image modeling, which models a sequence of quantized image tokens. VQGAN is a network that uses a feature quantization mechanism \cite{van2017neural} at the bottleneck of the encoder-decoder CNN to obtain the image tokens.

In our CompletionNets, \vqganOne{} encodes the incomplete image, and \vqganTwo{} decodes the features complemented by the transformer. Unlike TT, CompletionNets treats images of fixed size as input and output, considering that VQGAN also overfits the training resolution probably due to the inherent distortion of ERP representation, and uses WS-perceptual loss for training and circular inference for inference.
Our transformer models the scene of a 360-degree image as a sequence of quantized features and performs diverse image completions by sampling from the learned distribution.

\begin{figure*}
  \centering
  \vspace{-4mm}
  \begin{minipage}[t]{0.66\linewidth}
    \raisebox{8mm}{\includegraphics[width=\linewidth]{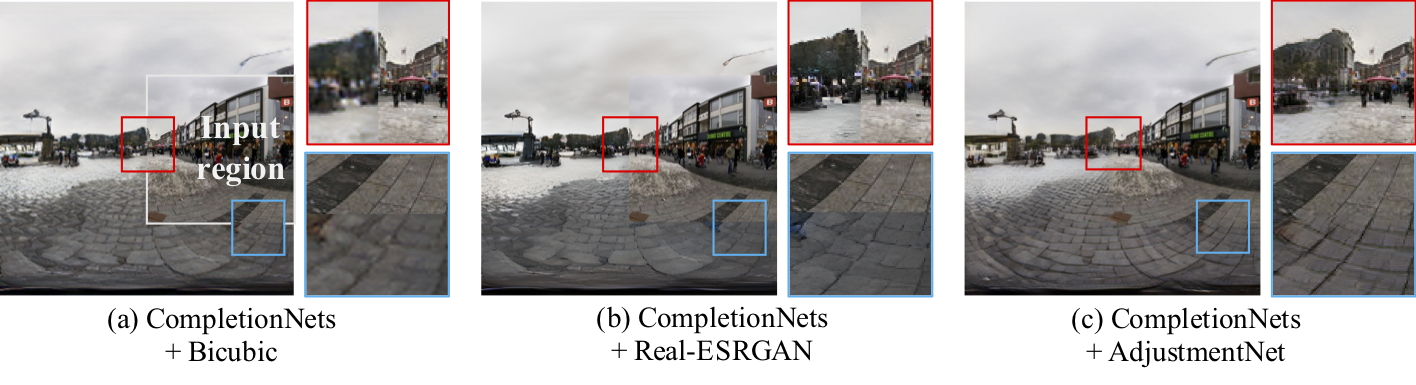}}
    \vspace{-12mm}
    \caption{Effect of AdjustmentNet. (a) and (b) show that the output of CompletionNets needs to be adjusted not only in resolution but also in color and stitching with the input region.}
    \label{fig:comp_adjustnet}
  \end{minipage}
  \begin{minipage}{0.1\linewidth}
  \end{minipage}
  \begin{minipage}[t]{0.33\linewidth}
    \includegraphics[width=\linewidth]{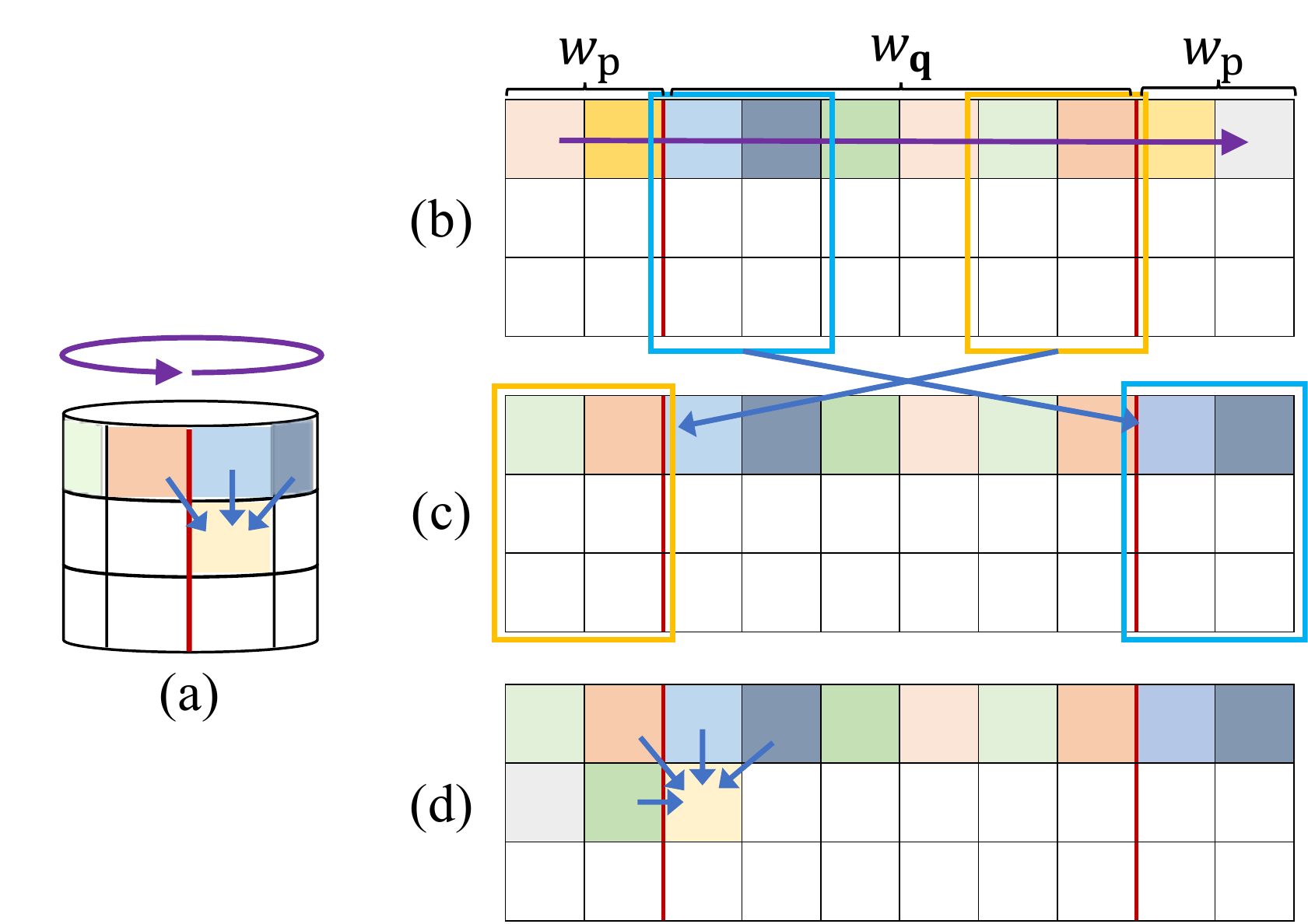}
    \caption{Circular inference.}
    \label{fig:circular}
  \end{minipage}
  \vspace{-4mm}
\end{figure*}

{\bf AdjustmentNet.} 
To achieve completion at arbitrary image sizes, we propose AdjustmentNet, a network that improves the consistency between the output of CompletionNets and the input region.
In high-resolution image completion methods employing two stages \cite{yi2020contextual, Wan_2021_ICCV}, the primary role of the second stage is to refine outputs by adding high-frequency components. However, as shown in Fig.~\ref{fig:comp_adjustnet}, applying only super-resolution is not sufficient for our method. Fig.~\ref{fig:comp_adjustnet}(a) shows a composite image of the upscaled completion image and the input region.  Fig.~\ref{fig:comp_adjustnet}(b) shows that even with the SOTA method of super-resolution \cite{wang2021realesrgan}, refinement alone is not sufficient. 
One of the causes for this is that the transformer resamples not only the completion region but also the input region. As a result, the completed region is predicted to fit the resampled input region and does not match the original input image.
In contrast, we adjust the output of CompletionNets to match the input image in terms of color, stitching, and resolution, as shown in (c). 
The network is a U-net structure implemented in the same CNN structure as VQGAN without the VQ mechanism \cite{van2017neural}.

\subsection{Training} \label{sec:training}
{\bf WS-perceptual loss.} VQGAN is a network for obtaining quantized vectors of image features, which models local regions of an image using CNN. TT proposes a self-supervised manner using adversarial loss $\mathcal{L}_\text{GAN}$, L1 loss $\mathcal{L}_\text{1}$, perceptual loss $\mathcal{L}_\text{Perc}$, and VQ loss $\mathcal{L}_\text{VQ}$. In contrast, we propose a novel loss function, WS-perceptual loss, to suitably model local regions for ERP representation. This loss function reflects the nature of ERP representation that there is a difference in the amount of information in each region along the latitudinal direction. Previous methods \cite{somanath2021hdr, gardner-sigasia-17} weighed pixel-level differential losses, such as L1 loss, to account for their projection onto a sphere. However, high-level features, such as semantics, should also be modeled around the central region. Therefore, WS-perceptual loss is an extension of perceptual loss (LPIPS) \cite{zhang2018perceptual} to the loss on the unit sphere as follows:
Similar to WS-PSNR \cite{sun2017weighted}, we prepare the following weights to account for the projection onto a sphere.\vspace{-2mm}
\begin{equation} \label{eq:weight}\vspace{-2mm}
    \small w'_l(u,v) = \cos ((v-H_{l}/2+1/2) \cdot \pi / H_{l}),
\end{equation}
where $u$ and $v$ are the positions on the feature (size $H_{l} \times W_{l}$) in the $l$th layer of the feature extractor.
We use Eq.~\ref{eq:weight} to weigh the perceptual loss $\small \mathcal{L}_\text{Perc} = \sum_{l} \frac{1}{H_l W_l} \sum_{u,v} \Vert w_l \odot (y^l_{uv} - x^l_{uv}) \Vert_2^2$ at each resolution.\vspace{-1mm}
\begin{equation}\vspace{-1mm}
    \small \mathcal{L}_\text{WS-Perc} = \sum_{l} \frac{1}{\sum_{u,v}w'_l} \sum_{u,v} w'_l \odot \Vert w_l \odot (y^l_{uv} - x^l_{uv}) \Vert_2^2.
\end{equation}

{\bf VQGAN.} We train both \vqganOne{} and \vqganTwo{}, which have both encoder and decoder, with\vspace{-2mm}
\begin{equation}
    \small \mathcal{L}_\text{VQGAN} = \lambda_\text{GAN} \mathcal{L}_\text{GAN} + \lambda_\text{1} \mathcal{L}_\text{1} + \lambda_\text{VQ} \mathcal{L}_\text{VQ} + \lambda_\text{WS-Perc} \mathcal{L}_\text{WS-Perc}.
\end{equation}
\vqganOne{} learns to reconstruct 360-degree images with missing regions to obtain quantized features $z_\quantize \in \RR^{h_{\quantize} \times w_{\quantize} \times n_{z}}$ of the incomplete input image. \vqganTwo{}, in contrast, learns to reconstruct a complete 360-degree image to obtain a decoder that obtains a complete 360-degree image from quantized features $\hat{z}_\quantize \in \RR^{h_{\quantize} \times w_{\quantize} \times n_{z}}$. 

{\bf Transformer.} We train the transformer to model a 360-degree scene and to perform completion. Using the transformation from $z_\quantize$ to $\hat{z}_\quantize$ as supervision, the model learns to predict the distribution of the next index after indices $s_{<i}$, conditional on indices $c$, by using the following equation:\vspace{-2mm}
\begin{equation}\vspace{-2mm}
  \small \mathcal{L}_{\text{Transformer}} = \mathbb{E}_{x \sim p(x)} \left[ -\log p(s|c) \right],
\end{equation}
where $p(s|c) = \prod_i p(s_i|s_{<i}, c)$, and the transformer does not directly deal with the quantized features ($z_\quantize$ and $\hat{z}_\quantize$) but treats the sequence of indices ($c$ and $s$) assigned to them. 

{\bf AdjustmentNet.} 
AdjustmentNet has a simple network architecture but explicitly learns to match the output image of CompletionNets with its input.
Therefore, we train AdjustmentNet by restoring the preprocessed input images to the original images (GT images) in a self-supervised manner. In addition, to avoid over-fitting to the top and bottom distortions of the ERP image, we randomly crop only a part from the ERP image and use it as the GT image. The following steps describe the preprocessing:
(1) To learn the adjustment of the boundary connection, reconstruct the GT image using the learned \vqganTwo{}, obtaining a reconstructed image with the difference from the GT image.
(2) To learn color adjustment, add color jitter to the input image before reconstructing it with \vqganTwo{}, which results in a reconstructed image with slightly different colors from the GT image.
(3) To learn to adjust the resolution, scale down the GT image to be reconstructed by \vqganTwo{} in advance. After reconstruction, the image is returned to the original scale using the bicubic method.
By composing this reconstructed image with a smaller region of the GT image than this image and using it as the input image for AdjustmentNet, we can learn to adjust color, stitching, and resolution while using the GT region as a hint. Therefore, we use the vanilla perceptual loss instead of the WS-perceptual loss.
We use \vspace{-2mm}
\begin{equation}\vspace{-2mm}
    \mathcal{L}_\text{Adjust} = \lambda_\text{GAN} \mathcal{L}_\text{GAN} + \lambda_\text{1} \mathcal{L}_\text{1} + \lambda_\text{Perc} \mathcal{L}_\text{Perc}
\end{equation}
for the learning.

\subsection{Inference} \label{sec:inference}
{\bf Circular inference.} 
To obtain completion results that reflect the continuity of 360-degree images, we propose circular inference as an inference order for the transformer.
TT proposes a raster ordering, called sliding attention window, as the autoregressive order of the transformer estimation. However, as shown in Fig.~\ref{fig:limitations_prior}(b), this method discontinues at both ends of the 360-degree image. SIG-SS proposes circular padding to connect the two ends. However, this padding is for convolution and cannot be applied to estimate the global semantics of 360-degree images using a transformer. Therefore, we propose circular inference so that the continuity of both ends can be accounted for in the estimation stage of the transformer, as shown in Fig.~\ref{fig:circular}(a) and Fig.~\ref{fig:circular}(d). The main idea is to circularly estimate some regions twice by the transformer to generate overlaps. 
For implementation, we duplicate both ends (with length $w_{\text{p}}$) of the quantized feature map $z_\quantize \in \RR^{h_{\quantize} \times w_{\quantize} \times n_{z}}$ on the opposite side in advance (from $h_{\quantize} \times w_{\quantize}$ to $h_{\quantize} \times (w_{\quantize}+2w_{\text{p}}$)), and the transformer performs an estimation in raster order (Fig.~\ref{fig:circular}(b)). After estimating a row, the estimation results of both ends of the quantized feature map $\hat{z}_\quantize \in \RR^{h_{\quantize} \times w_{\quantize} \times n_{z}}$ are copied from the opposite side and replaced (Fig.~\ref{fig:circular}(c)). That is, the length $w_{\text{p}}$ from the left is replaced by the estimated result of $w_{\text{p}}$ from the right of the original length $w_{\quantize}$. The same applies to the other side.

As described above, circularly estimating with a transformer allows connecting the two ends in higher-order features. When decoded into an image, the connection at the semantics level is improved, allowing for a more plausible completion as a 360-degree image. 

Thus, the overall inference is \vspace{-2mm}
\begin{equation}\vspace{-2mm}
    y = G_{\text{Adjust}}(G_{\text{\vqganTwo{}}}(T(E_{\text{\vqganOne{}}}(x'))), x'),
\end{equation}
where $G_{\text{Adjust}}$, $G_{\text{\vqganTwo{}}}$, $T$, and $E_{\text{\vqganOne{}}}$ indicate AdjustmentNet, the decoder of \vqganTwo{}, the transformer, and the encoder of \vqganOne{}, respectively.

\begin{figure*}[t]
  \centering
  \vspace{-1mm}
   \includegraphics[width=0.94\linewidth]{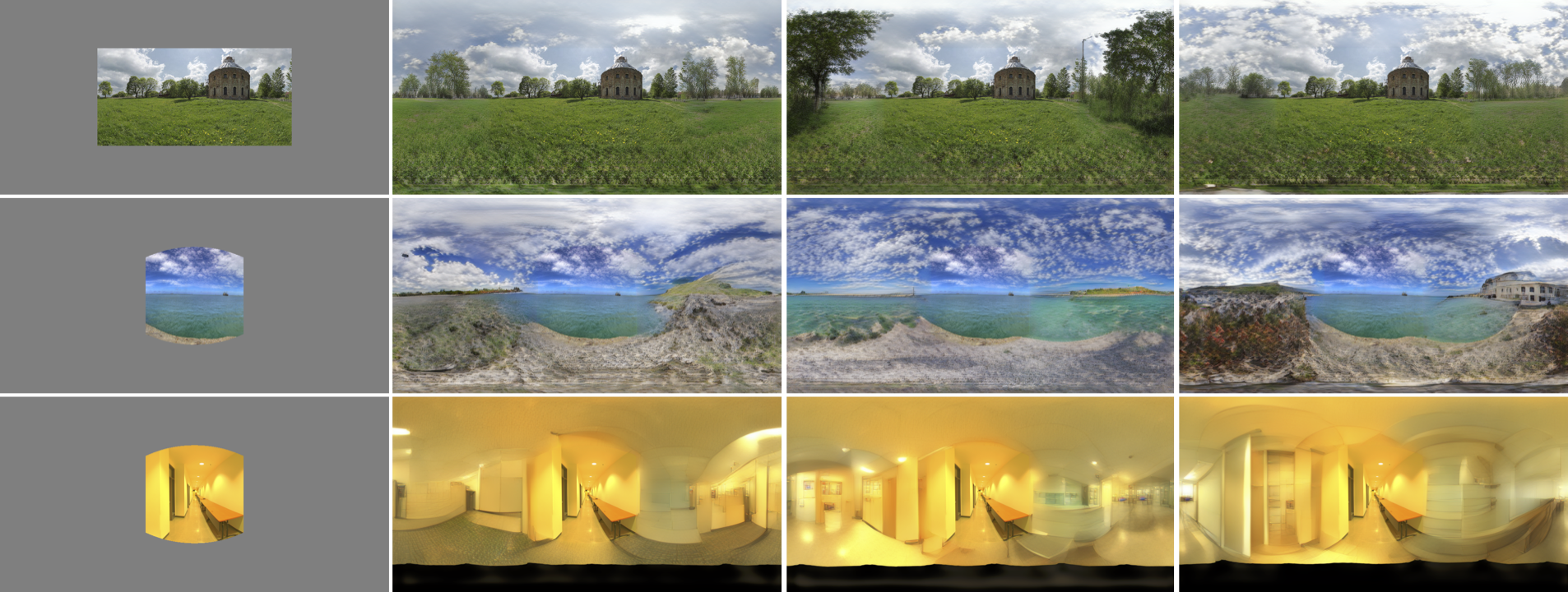}
   \caption{Diverse outputs of the proposed method. Our method performs diverse and plausible completions on a given input (the first column).}
   \label{fig:diverse}
   \vspace{-4mm}
\end{figure*}

\section{Experiments}
\label{sec:experiments}
We compare our method qualitatively and quantitatively with previous methods to verify the effectiveness of the proposed components.
In our supplementary video, we also compare the results in all-around viewing and object insertion applications.

\subsection{Experiment Settings}
{\bf Implementation details.} 
We use Adam \cite{kingma2014adam} as an optimizer whose learning rate = 4.5e-06. $\lambda_\text{1} = 1.0$, $\lambda_\text{Perc} = 1.0$, $\lambda_\text{VQ} = 1.0$, $\lambda_\text{WS-Perc} = 1.0$, and $\lambda_\text{GAN}$ is an adaptive weight \cite{Esser_2021_CVPR}.
We train the transformer with 20 epochs and, for the remaining networks, with 30 epochs. Our method produces \Lsize{} images.

{\bf Datasets.}
We use SUN360 \cite{sun360} and Laval Indoor Dataset \cite{gardner-sigasia-17} as the datasets.
We divide SUN360 into 47938 training images and 5000 test images of the ``Outdoor'' class. In contrast, we divide the Laval Indoor Dataset into 1837 training images and 289 test images, following the provided training and test split. 
The number of images in the Laval Indoor Dataset is too tiny to train our model, so we fine-tune the model trained by SUN360 on this dataset, just as EnvMapNet uses an additional dataset.
For data augmentation, we randomly change the view direction horizontally, as in \cite{akimoto2019360, hara2021spherical}.

{\bf Evaluation.}
As a quantitative metric, we use Fréchet inception distance (FID) \cite{heusel2017gans} to evaluate the quality and diversity of the generated images. We explain the details of the calculation method in Sec. \ref{sec:quantitative}.
When comparing the proposed method with other methods, we set the sequence length of the transformer to 512.
When validating each component of the proposed method, we set the sequence length of the transformer to 256 to make network training more efficient.

{\bf Baselines.}
To compare our model with 360IC, we train 360IC on our dataset. The original 360IC introduces a two-stage approach to avoid overtraining on a small training dataset of 600 images. We implement 360IC with a single-stage as in \cite{hara2021spherical}, and then we train the model with a sufficient amount of training data. To compare the differences in scene modeling between a transformer and CNN, we use a CNN structure (Fig.~\ref{fig:limitations_prior}(a)) similar to that of VQGAN and the same training procedure as ours. Furthermore, we implement the bottleneck by adding their proposed four parallel dilated conv blocks instead of the quantization mechanism. 

To compare with SIG-SS, we infer the authors' trained models with our test images. The authors trained their model on SUN360. Note, however, that their train/test splitting is unknown, and they may have trained their model on the test images we prepared.

EnvMapNet does not publish their code, only their evaluation protocol and scripts. We follow their instructions and run their evaluation script on the same train/test split of the Laval Indoor Dataset. For comparison, we quote the resulting images and scores from their paper. Note that the location of the input region and the tone mapping method do not match their experiment and ours.

\subsection{Diverse Outputs}
Fig.~\ref{fig:diverse} shows that our approach can output multiple and diverse completions. The left column is an input image, and all results are \Lsize{} images. The top two rows show the results generated with the same model trained with SUN360 and different input regions. In the top row, the input regions are 180 degrees in the longitudinal direction and 90 degrees in the latitudinal direction. The middle row input regions correspond to the 90-degree angle of view when converted to a perspective image. We can find that the larger the input region, the better the quality of the generated texture.
The last row shows the results of an experiment conducted using the Laval Indoor Dataset and that our method can estimate indoor scenes with various structures.

\subsection{Qualitative Comparisons} \label{sec:qualitative}
\begin{figure}[t]
  \centering
  \vspace{-1mm}
   \includegraphics[width=\linewidth]{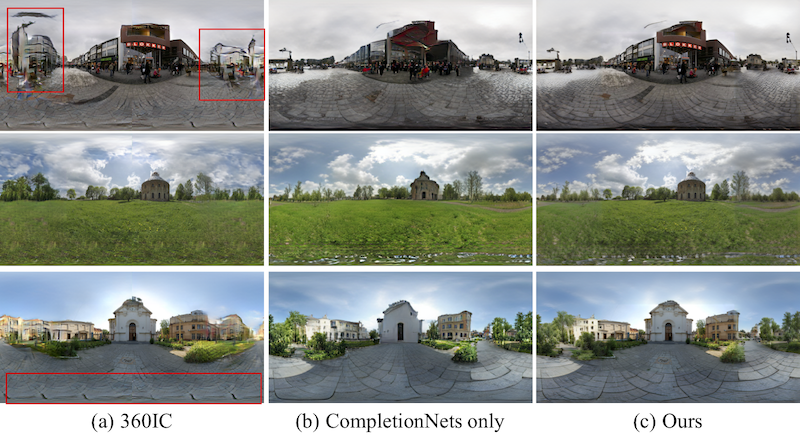}
   \caption{Qualitative comparison with 360IC. The input region is the same as that in the first row of Fig.~\ref{fig:diverse}.}
   \label{fig:comp_360ic}
  \centering
  \vspace{0.5mm}
   \includegraphics[width=\linewidth]{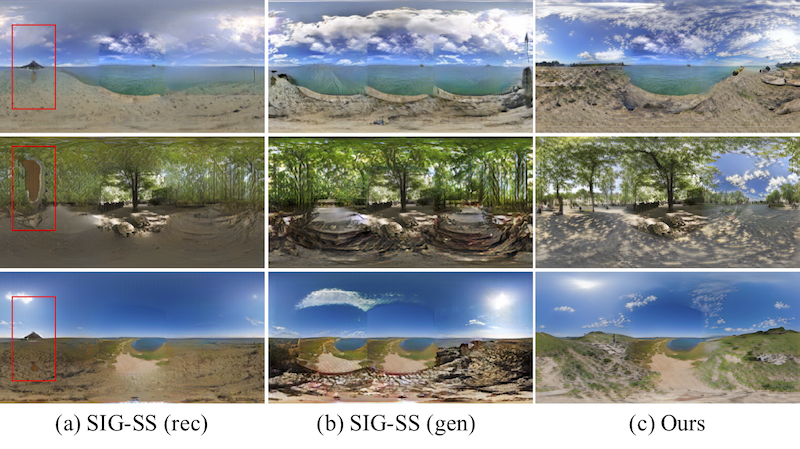}
   \caption{Qualitative comparison with SIG-SS. The input region is the same as that in the second row of Fig.~\ref{fig:diverse}.}
   \label{fig:comp_sigss}
  \centering
  \vspace{0.5mm}
   \includegraphics[width=\linewidth]{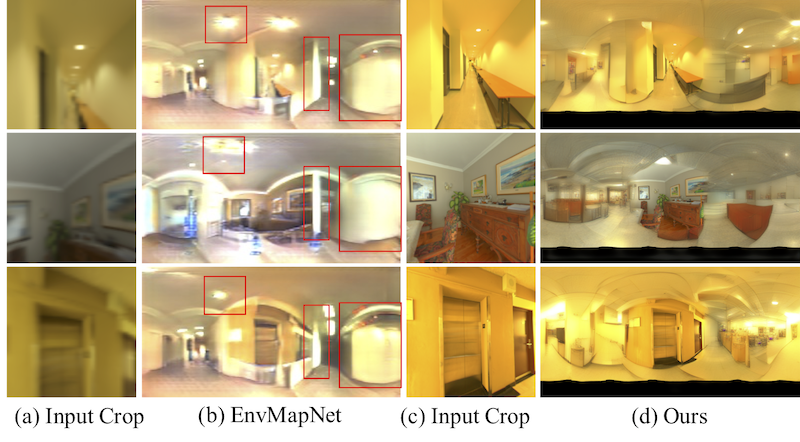}
   \caption{Qualitative comparison with EnvMapNet. The input area is the same as that in the third row of Fig.~\ref{fig:diverse}. Note that the inputs of EnvMapNet(a) and ours(c) are not exactly the same.}
   \label{fig:comp_envmapnet}
\end{figure}

Fig.~\ref{fig:comp_360ic} compares 360IC and the proposed method. 360IC captures the distortions of ERP representation, but there are artifacts in the textures.
In contrast, the proposed method can generate the textures and shapes of each object more accurately. 
Comparing ``360IC'' and ``CompNets Only,'' we can see the difference in scene modeling between convolution and transformer. Dilated convolution increases the receptive field of the CNN, but this causes the transferred information to be sparse, which may be the cause of artifacts in the texture generation. In contrast, the transformer can generate globally consistent textures and represent distortions in the upper and lower regions of the image, where distortions specific to ERP images occur significantly.

Fig.~\ref{fig:comp_sigss} compares SIG-SS and ours. The input region is the same as in Fig.~\ref{fig:diverse}(b); SIG-SS has the results of reconstruction (rec) and sampling (gen). The resolution is \Msize. The reconstruction results show overfitting, where similar objects appear, and the sampling results lack global consistency.
In contrast, the proposed method provides results that match the context of the input region.

Fig.~\ref{fig:comp_envmapnet} shows a comparison between ours and EnvMapNet, where EnvMapNet is a \Ssize{} completion image, while ours is a \Lsize{} completion image. In other words, our method can complete 16 times as many pixels as EnvMapNet, and our results are also better looking. The preprocessing of EnvMapNet, which requires clustering of datasets to stabilize adversarial learning, is not necessary for our method. The results trained on the Laval Indoor Dataset do not generate detailed textures compared to those trained on SUN360. This is because we train the model on a small dataset, which is one of our limitations.


\subsection{Quantitative Comparisons} \label{sec:quantitative}
\vspace{-1mm}
We use FID to evaluate the quality and diversity of the completions of the proposed method. 
Table \ref{table:comp_360ic} uses SUN360 and Clean-FID \cite{parmar2021cleanfid} as the script to compute the FID. The results show that our method outperforms 360IC. 
Using the same evaluation method, in Table \ref{table:comp_sig-ss}, we compare our method with SIG-SS, showing that our method, which uses a transformer, outperforms their method, which uses CVAE.
Table \ref{table:comp_laval} compares our method with EnvMapNet and that of Gardner et al. \cite{gardner-sigasia-17} on the Laval Indoor Dataset. To compute the FID, we follow their evaluation protocol: we convert an image into a Cubemap and remove the top and bottom planes containing little information. 
In summary, the FID comparisons show that our method is superior in terms of the generated results' quality and diversity.

\begin{figure}[t]
  \centering
  \vspace{-1mm}
   \includegraphics[width=\linewidth]{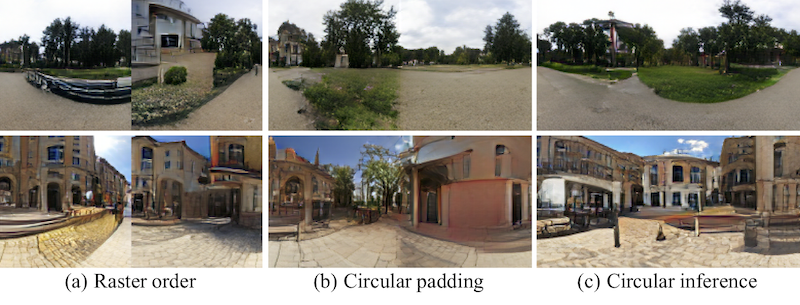}
   \caption{Circular inference connects both ends of a $360^{\circ}$ image, both at the pixel and semantic levels. See Sec. \ref{sec:analysis} for details.}
   \label{fig:effect_circular_inference}
\end{figure}

\begin{table*}[t]
  \centering
  \vspace{-1mm}
  \begin{minipage}[t]{0.33\linewidth}
    \centering
    \resizebox{0.99\textwidth}{!}{
    \begin{tabular}{lccc}
      \toprule
      & 360IC\cite{akimoto2019360} & CompletionNets Only & Ours \\
      \midrule
      FID$\downarrow$ & 16.44 & 14.96 & {\bf 9.52} \\
      \bottomrule
    \end{tabular}
    } 
  \caption{\small FID on SUN360 w/ $180^{\circ}\times90^{\circ}$ input. } 
  \label{table:comp_360ic} 
  \end{minipage}
  \begin{minipage}[t]{0.33\linewidth}
    \centering
    \resizebox{0.99\textwidth}{!}{
    \begin{tabular}{lccc}
      \toprule
      & SIG-SS(rec)\cite{hara2021spherical} & SIG-SS(gen)\cite{hara2021spherical} & Ours \\
      \midrule
      FID$\downarrow$ & 31.91 & 26.81 & {\bf 23.13} \\
      \bottomrule
    \end{tabular}
    } 
  \caption{\small FID on SUN360 w/ $90^{\circ}$ input.} 
  \label{table:comp_sig-ss} 
  \end{minipage}
  \begin{minipage}[t]{0.33\linewidth}
    \centering
    \resizebox{0.99\textwidth}{!}{
    \begin{tabular}{lccc}
      \toprule
      & Gardner {\it et al.} \cite{gardner-sigasia-17} & EnvMapNet\cite{somanath2021hdr} & Ours \\
      \midrule
      FID$\downarrow$ & 197.4 & 52.7 & {\bf 46.15} \\
      \bottomrule
    \end{tabular}
    } 
  \caption{\small FID on Laval Indoor dataset. } 
  \label{table:comp_laval} 
  \end{minipage}

\end{table*}
\begin{table*}[t]
  \centering
  \vspace{-2mm}
  \begin{minipage}[t]{0.33\linewidth}
    \centering
    \resizebox{0.99\textwidth}{!}{
    \begin{tabular}{lccc}
      \toprule
      & Raster order & Circular padding & Circular inference \\
      \midrule
      FID$\downarrow$ & 30.03 & {\bf 26.33} & 26.96 \\
      \bottomrule
    \end{tabular}
    } 
  \caption{\footnotesize Evaluation of circular inference on SUN360 w/ $180^{\circ}\times90^{\circ}$ input. } 
  \label{table:circular} 
  \end{minipage}
  \begin{minipage}[t]{0.33\linewidth}
    \centering
    \resizebox{0.99\textwidth}{!}{
    \begin{tabular}{lccc}
      \toprule
      & Perceptual loss & WS-L1 loss & WS-perceptual loss \\
      \midrule
      FID$\downarrow$ & 29.00 & 35.00 & {\bf 26.96} \\
      \bottomrule
    \end{tabular}
    } 
  \caption{\footnotesize Evaluation of WS-perceptual loss on the proposed network on SUN360 w/ $180^{\circ}\times90^{\circ}$ input.} 
  \label{table:ws-percptual} 
  \end{minipage}
  \begin{minipage}[t]{0.33\linewidth}
    \centering
    \resizebox{0.80\textwidth}{!}{
    \begin{tabular}{lcc}
      \toprule
      & Perceptual loss &  WS-perceptual loss \\
      \midrule
      FID$\downarrow$ & 67.47 & {\bf 50.87} \\
      \bottomrule
    \end{tabular}
    } 
  \caption{\footnotesize Evaluation of WS-perceptual loss on the 360IC network on SUN360 w/ $180^{\circ}\times90^{\circ}$ input. } 
  \label{table:ws-icip} 
  \end{minipage}
\vspace{-4mm}
\end{table*}

\subsection{Analysis} \label{sec:analysis}
{\bf Verification of circular inference.}
Fig.~\ref{fig:effect_circular_inference} and Table \ref{table:circular} show that our circular inference can perform consistent estimation as ERP images. In Fig.~\ref{fig:effect_circular_inference}, we show a partially cropped region (\Ssize{}) of the output of CompletionNet (\Msize{}), with the image ends aligned such that the completion region is in the image center. We use the same trained model for each method and change only some of the inferences. As shown in Fig.~\ref{fig:effect_circular_inference}(a), the left and right edges are not connected in the raster order, which is the auto-regression of a transformer used in TT. One way to connect the two edges is to use circular padding \cite{hara2021spherical}. Circular padding is a technique that pads the pixels on opposite edges of the ERP image during convolution. We use it to decode the features estimated by our transformer into images. This technique helps to improve the continuity at the pixel level, but at the semantic level, the contents at both ends are different. For example, in Fig.~\ref{fig:effect_circular_inference}(b), the grass and dirt ground are separated in the center. In contrast, circular inference helps to improve the continuity at the semantic level during transformer estimation. As Table \ref{table:circular} shows, the FID scores of circular padding and circular inference are similar, but qualitatively, circular inference contributes to the generation of images that better reflect ERP image properties.

{\bf Verification of WS-perceptual loss.}
The loss considers the difference in pixel level and the difference in high-level features in the information amount along the latitudinal direction of the ERP image. In Table \ref{table:ws-percptual} and Table \ref{table:ws-icip}, to evaluate the effect more directly, we do not use AdjustmentNet but the output of CompletionNets. To compute the FID only in the completed region and not in the input region, we use only the generated region (\Ssize{}) in the image center corresponding to 90 degrees of latitude and 180 degrees of longitude out of the output image (\Msize{}). Table \ref{table:ws-percptual} compares the results of WS-perceptual loss with perceptual loss or with WS-L1 loss, which considers only low-level differences, on the proposed network. Table \ref{table:ws-icip} shows that the use of WS-perceptual loss contributes to improving the FID score in training the 360IC network. In summary, WS-perceptual loss contributes to generating images with more ERP image properties by considering the sphere and weighing high-level features.

\section{Application}
\label{sec:application}
\begin{figure}[t]
  \centering
  \vspace{-2mm}
   \includegraphics[width=1.0\linewidth]{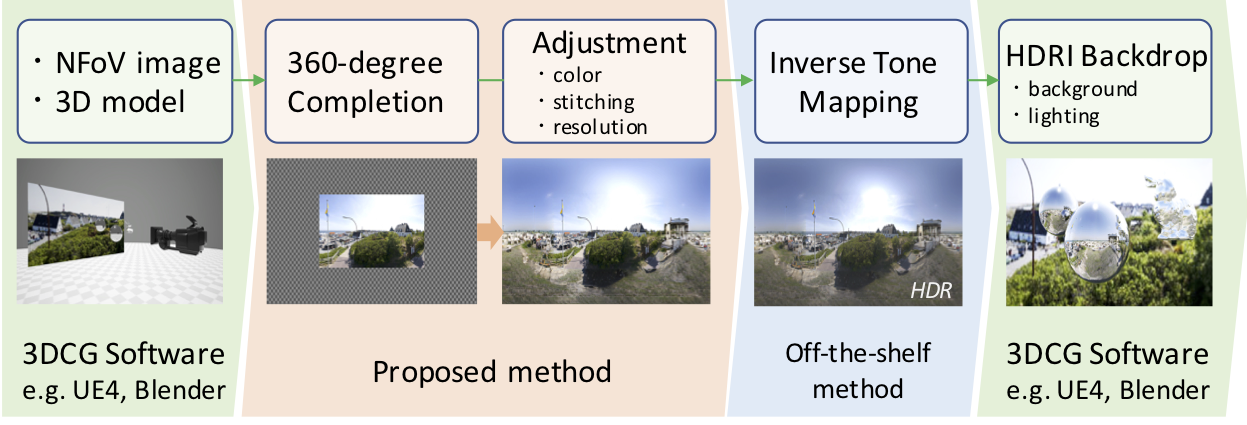}
   \caption{Pipeline for using the completion results for lighting and background of 3DCG scenes.}
   \label{fig:backdrop_pipeline}
\end{figure}

We implement an application that uses a completed 360-degree image as a background and for lighting. We propose a pipeline to realize image-based lighting on the 3DCG software Unreal Engine 4 (UE4), as shown in Fig.~\ref{fig:backdrop_pipeline}. To use an HDRI Backdrop plugin, we need to convert images to HDR images beforehand; we convert low dynamic range (LDR) to high dynamic range (HDR) by inverse tone mapping using existing methods \cite{liu2020single}. By introducing this process, our method can generate HDR environment maps from an LDR NFoV image.

Using this pipeline, we demonstrate the background creation and lighting for specular objects placed in a 3DCG scene in the bottom row of Fig.~\ref{fig:teaser} and the supplementary video. 
Because of our proposed method's high resolution and plausible 360-degree image completion, we can demonstrate moving a camera to look all around. To the best of our knowledge, no other work has achieved such demonstration. When an NFoV image is composited behind 3DCG objects, the specular surfaces cannot be represented. Previous works \cite{somanath2021hdr,wang2021learning}, which have often been limited to indoor environments, have generated environment maps to represent specular reflections; however, they have much lower resolution than our method. Therefore, there is no other method to insert specular objects as close to the camera as our results.

\vspace{-1.2mm}
\section{Discussion}
\label{sec:discussion}
This paper addresses the problem of completing 360-degree images from a narrow field of view. We first reveal that the previous methods had the limitations of overfitting the training resolution and having deterministic outputs. Next, to propose a framework to solve these problems, we introduce a transformer-based diverse Im2Im; however, the resolution problem remains. Thus, we propose AdjustmentNet. Moreover, we propose two novel techniques for obtaining completion results with improved properties of 360-degree images. Finally, our demonstrations show that, unlike others, the proposed method can provide designers with a new workflow for 3DCG creation by efficiently delivering an all-surrounding background.

\subsection{Limitation}
{\bf Inference time and computational memory.} Our method takes approximately 30 s for one completion on a single 2080Ti, mainly because of a transformer. However, since 3DCG designers, the subject of our work, are likely to use PCs and servers with high-end GPUs, the performance of our method may be sufficient for practical use. 

{\bf Controllability.} The proposed method does not control what is generated in the completion region. One possible solution is to paste an object that appears directly in the completion region and complete it so that it is smoothly connected.

\subsection{Potential Impact}
This method is helpful for photo-based background representations in {\bf virtual production}. For example, to project a background on an LED wall, designers occasionally create a background by compositing images obtained from stock photo sites instead of using a 3D environment composed of 3D models. However, it is possible to estimate the situations in areas the images do not photograph and express reflections on an inserted object.

Due to the limited number of photos available on {\bf HDRI's stock photo} site, designers often lack originality as they use the same images. Even a popular website \cite{polyhaven} provides only approximately 490 images, which is very few compared to our 5000 test results. Our method solves this problem by generating various new 360-degree HDRIs. 

By streaming only the area the user is looking at out of the 360 degrees, we can reduce the communication capacity of video streaming in virtual reality spaces, such as {\bf the Metaverse}. However, when the end-user's avatar holds a mirror in its hand, it may not reflect anything. In this case, our method can render the estimated scene.

{\bf Negative impact}. Our completion results have room for improvement compared to real 360-degree images. However, if the generation results are improved, it could be a kind of Deepfake. Realistic generation of non-existent scenes or compositing of fake objects can mislead people.


{\small
\bibliographystyle{ieee_fullname}
\bibliography{egbib}

\begin{thebibliography}{10}\itemsep=-1pt

\bibitem{akimoto2019360}
Naofumi Akimoto, Seito Kasai, Masaki Hayashi, and Yoshimitsu Aoki.
\newblock 360-degree image completion by two-stage conditional gans.
\newblock In {\em 2019 IEEE International Conference on Image Processing
  (ICIP)}, pages 4704--4708. IEEE, 2019.

\bibitem{barnes2009patchmatch}
Connelly Barnes, Eli Shechtman, Adam Finkelstein, and Dan~B Goldman.
\newblock Patchmatch: A randomized correspondence algorithm for structural
  image editing.
\newblock {\em ACM Trans. Graph.}, 28(3):24, 2009.

\bibitem{chen2020generative}
Mark Chen, Alec Radford, Rewon Child, Jeffrey Wu, Heewoo Jun, David Luan, and
  Ilya Sutskever.
\newblock Generative pretraining from pixels.
\newblock In {\em International Conference on Machine Learning}, pages
  1691--1703. PMLR, 2020.

\bibitem{Esser_2021_CVPR}
Patrick Esser, Robin Rombach, and Bjorn Ommer.
\newblock Taming transformers for high-resolution image synthesis.
\newblock In {\em Proceedings of the IEEE/CVF Conference on Computer Vision and
  Pattern Recognition (CVPR)}, pages 12873--12883, June 2021.

\bibitem{Gardner_2019_ICCV}
Marc-Andre Gardner, Yannick Hold-Geoffroy, Kalyan Sunkavalli, Christian Gagne,
  and Jean-Francois Lalonde.
\newblock Deep parametric indoor lighting estimation.
\newblock In {\em Proceedings of the IEEE/CVF International Conference on
  Computer Vision (ICCV)}, 2019.

\bibitem{gardner-sigasia-17}
Marc-Andr\'{e} Gardner, Kalyan Sunkavalli, Ersin Yumer, Xiaohui Shen, Emiliano
  Gambaretto, Christian Gagn\'{e}, and Jean-Fran\c{c}ois Lalonde.
\newblock Learning to predict indoor illumination from a single image.
\newblock {\em ACM Transactions on Graphics (SIGGRAPH Asia)}, 2017.

\bibitem{hara2021spherical}
Takayuki Hara, Yusuke Mukuta, and Tatsuya Harada.
\newblock Spherical image generation from a single image by considering scene
  symmetry.
\newblock In {\em Proceedings of the AAAI Conference on Artificial
  Intelligence}, 2021.

\bibitem{Hays2007scene}
James Hays and Alexei~A Efros.
\newblock Scene completion using millions of photographs.
\newblock {\em ACM Transactions on Graphics (SIGGRAPH 2007)}, 26(3), 2007.

\bibitem{heusel2017gans}
Martin Heusel, Hubert Ramsauer, Thomas Unterthiner, Bernhard Nessler, and Sepp
  Hochreiter.
\newblock Gans trained by a two time-scale update rule converge to a local nash
  equilibrium.
\newblock {\em Advances in neural information processing systems}, 30, 2017.

\bibitem{IizukaSIGGRAPH2017}
Satoshi Iizuka, Edgar Simo-Serra, and Hiroshi Ishikawa.
\newblock {Globally and Locally Consistent Image Completion}.
\newblock {\em ACM Transactions on Graphics (Proc. of SIGGRAPH 2017)}, 2017.

\bibitem{kingma2014adam}
Diederik~P Kingma and Jimmy Ba.
\newblock Adam: A method for stochastic optimization.
\newblock {\em arXiv preprint arXiv:1412.6980}, 2014.

\bibitem{legendre2019deeplight}
Chloe LeGendre, Wan-Chun Ma, Graham Fyffe, John Flynn, Laurent Charbonnel, Jay
  Busch, and Paul Debevec.
\newblock Deeplight: Learning illumination for unconstrained mobile mixed
  reality.
\newblock In {\em Proceedings of the IEEE/CVF Conference on Computer Vision and
  Pattern Recognition}, pages 5918--5928, 2019.

\bibitem{li2017generative}
Yijun Li, Sifei Liu, Jimei Yang, and Ming-Hsuan Yang.
\newblock Generative face completion.
\newblock In {\em Proceedings of the IEEE conference on computer vision and
  pattern recognition}, pages 3911--3919, 2017.

\bibitem{lin2019coco}
Chieh~Hubert Lin, Chia-Che Chang, Yu-Sheng Chen, Da-Cheng Juan, Wei Wei, and
  Hwann-Tzong Chen.
\newblock Coco-gan: Generation by parts via conditional coordinating.
\newblock In {\em Proceedings of the IEEE/CVF International Conference on
  Computer Vision}, pages 4512--4521, 2019.

\bibitem{liu2021infinite}
Andrew Liu, Richard Tucker, Varun Jampani, Ameesh Makadia, Noah Snavely, and
  Angjoo Kanazawa.
\newblock Infinite nature: Perpetual view generation of natural scenes from a
  single image.
\newblock In {\em Proceedings of the IEEE/CVF International Conference on
  Computer Vision}, pages 14458--14467, 2021.

\bibitem{liu2019coherent}
Hongyu Liu, Bin Jiang, Yi Xiao, and Chao Yang.
\newblock Coherent semantic attention for image inpainting.
\newblock In {\em Proceedings of the IEEE/CVF International Conference on
  Computer Vision}, pages 4170--4179, 2019.

\bibitem{liu2020single}
Yu-Lun Liu, Wei-Sheng Lai, Yu-Sheng Chen, Yi-Lung Kao, Ming-Hsuan Yang, Yung-Yu
  Chuang, and Jia-Bin Huang.
\newblock Single-image hdr reconstruction by learning to reverse the camera
  pipeline.
\newblock In {\em Proceedings of the IEEE/CVF Conference on Computer Vision and
  Pattern Recognition}, pages 1651--1660, 2020.

\bibitem{parmar2021cleanfid}
Gaurav Parmar, Richard Zhang, and Jun-Yan Zhu.
\newblock On buggy resizing libraries and surprising subtleties in fid
  calculation.
\newblock {\em arXiv preprint arXiv:2104.11222}, 2021.

\bibitem{pathak2016context}
Deepak Pathak, Philipp Krahenbuhl, Jeff Donahue, Trevor Darrell, and Alexei~A
  Efros.
\newblock Context encoders: Feature learning by inpainting.
\newblock In {\em Proceedings of the IEEE conference on computer vision and
  pattern recognition}, pages 2536--2544, 2016.

\bibitem{rombach2021geometryfree}
Robin Rombach, Patrick Esser, and Björn Ommer.
\newblock Geometry-free view synthesis: Transformers and no 3d priors, 2021.

\bibitem{neuralSengupta19}
Soumyadip Sengupta, Jinwei Gu, Kihwan Kim, Guilin Liu, David~W. Jacobs, and Jan
  Kautz.
\newblock Neural inverse rendering of an indoor scene from a single image.
\newblock In {\em International Conference on Computer Vision (ICCV)}, 2019.

\bibitem{Skorokhodov_2021_ICCV}
Ivan Skorokhodov, Grigorii Sotnikov, and Mohamed Elhoseiny.
\newblock Aligning latent and image spaces to connect the unconnectable.
\newblock In {\em Proceedings of the IEEE/CVF International Conference on
  Computer Vision (ICCV)}, pages 14144--14153, October 2021.

\bibitem{sohn2015learning}
Kihyuk Sohn, Honglak Lee, and Xinchen Yan.
\newblock Learning structured output representation using deep conditional
  generative models.
\newblock {\em Advances in neural information processing systems},
  28:3483--3491, 2015.

\bibitem{somanath2021hdr}
Gowri Somanath and Daniel Kurz.
\newblock Hdr environment map estimation for real-time augmented reality.
\newblock In {\em Proceedings of the IEEE/CVF Conference on Computer Vision and
  Pattern Recognition}, pages 11298--11306, 2021.

\bibitem{song2019neural}
Shuran Song and Thomas Funkhouser.
\newblock Neural illumination: Lighting prediction for indoor environments.
\newblock {\em Proceedings of 33th IEEE Conference on Computer Vision and
  Pattern Recognition}, 2019.

\bibitem{sun2017weighted}
Yule Sun, Ang Lu, and Lu Yu.
\newblock Weighted-to-spherically-uniform quality evaluation for
  omnidirectional video.
\newblock {\em IEEE signal processing letters}, 24(9):1408--1412, 2017.

\bibitem{teterwak2019boundless}
Piotr Teterwak, Aaron Sarna, Dilip Krishnan, Aaron Maschinot, David Belanger,
  Ce Liu, and William~T Freeman.
\newblock Boundless: Generative adversarial networks for image extension.
\newblock In {\em Proceedings of the IEEE/CVF International Conference on
  Computer Vision}, pages 10521--10530, 2019.

\bibitem{van2017neural}
A{\"a}ron van~den Oord, Oriol Vinyals, and Koray Kavukcuoglu.
\newblock Neural discrete representation learning.
\newblock In {\em NIPS}, 2017.

\bibitem{vaswani2017attention}
Ashish Vaswani, Noam Shazeer, Niki Parmar, Jakob Uszkoreit, Llion Jones,
  Aidan~N Gomez, {\L}ukasz Kaiser, and Illia Polosukhin.
\newblock Attention is all you need.
\newblock In {\em Advances in neural information processing systems}, pages
  5998--6008, 2017.

\bibitem{Wan_2021_ICCV}
Ziyu Wan, Jingbo Zhang, Dongdong Chen, and Jing Liao.
\newblock High-fidelity pluralistic image completion with transformers.
\newblock In {\em Proceedings of the IEEE/CVF International Conference on
  Computer Vision (ICCV)}, pages 4692--4701, October 2021.

\bibitem{wang2021realesrgan}
Xintao Wang, Liangbin Xie, Chao Dong, and Ying Shan.
\newblock Real-esrgan: Training real-world blind super-resolution with pure
  synthetic data.
\newblock In {\em International Conference on Computer Vision Workshops
  (ICCVW)}, 2021.

\bibitem{wang2019wide}
Yi Wang, Xin Tao, Xiaoyong Shen, and Jiaya Jia.
\newblock Wide-context semantic image extrapolation.
\newblock In {\em Proceedings of the IEEE/CVF Conference on Computer Vision and
  Pattern Recognition}, pages 1399--1408, 2019.

\bibitem{wang2021learning}
Zian Wang, Jonah Philion, Sanja Fidler, and Jan Kautz.
\newblock Learning indoor inverse rendering with 3d spatially-varying lighting.
\newblock In {\em Proceedings of International Conference on Computer Vision
  (ICCV)}, 2021.

\bibitem{wiles2020synsin}
Olivia Wiles, Georgia Gkioxari, Richard Szeliski, and Justin Johnson.
\newblock Synsin: End-to-end view synthesis from a single image.
\newblock In {\em Proceedings of the IEEE/CVF Conference on Computer Vision and
  Pattern Recognition}, pages 7467--7477, 2020.

\bibitem{sun360}
Jianxiong Xiao, Krista~A. Ehinger, Aude Oliva, and Antonio Torralba.
\newblock Recognizing scene viewpoint using panoramic place representation.
\newblock In {\em 2012 IEEE Conference on Computer Vision and Pattern
  Recognition}, pages 2695--2702, 2012.

\bibitem{yang2019very}
Zongxin Yang, Jian Dong, Ping Liu, Yi Yang, and Shuicheng Yan.
\newblock Very long natural scenery image prediction by outpainting.
\newblock In {\em Proceedings of the IEEE International Conference on Computer
  Vision}, pages 10561--10570, 2019.

\bibitem{yi2020contextual}
Zili Yi, Qiang Tang, Shekoofeh Azizi, Daesik Jang, and Zhan Xu.
\newblock Contextual residual aggregation for ultra high-resolution image
  inpainting.
\newblock In {\em Proceedings of the IEEE/CVF Conference on Computer Vision and
  Pattern Recognition}, pages 7508--7517, 2020.

\bibitem{yu2018generative}
Jiahui Yu, Zhe Lin, Jimei Yang, Xiaohui Shen, Xin Lu, and Thomas~S Huang.
\newblock Generative image inpainting with contextual attention.
\newblock In {\em Proceedings of the IEEE conference on computer vision and
  pattern recognition}, pages 5505--5514, 2018.

\bibitem{polyhaven}
Greg Zaal.
\newblock {\em Poly Haven}, 2021.

\bibitem{zhang2018perceptual}
Richard Zhang, Phillip Isola, Alexei~A Efros, Eli Shechtman, and Oliver Wang.
\newblock The unreasonable effectiveness of deep features as a perceptual
  metric.
\newblock In {\em CVPR}, 2018.

\bibitem{zheng2019pluralistic}
Chuanxia Zheng, Tat-Jen Cham, and Jianfei Cai.
\newblock Pluralistic image completion.
\newblock In {\em Proceedings of the IEEE Conference on Computer Vision and
  Pattern Recognition}, pages 1438--1447, 2019.

\end{thebibliography}
}

\clearpage
\appendix
\section{Additional Results}
\label{sec:additional}
In this section, we show figures of results that could not be included in the main body of the paper due to space constraints. Each result image is a higher-resolution image and is best viewed in color and enlarged on a display. Fig.~\ref{fig:supp_diverse} shows diverse completion results from our method, corresponding to Fig. \ref{fig:diverse}. Furthermore, Fig.~\ref{fig:supp_vs360ic}, \ref{fig:supp_vssigss} and \ref{fig:supp_vsenvmapnet} show comparisons with previous methods, corresponding to Fig.~\ref{fig:comp_360ic}, \ref{fig:comp_sigss} and \ref{fig:comp_envmapnet}.
\section{Additional Analyses}
\subsection{Generated Textures}
\begin{figure*}[t]
    \centering
    \includegraphics[width=\linewidth]{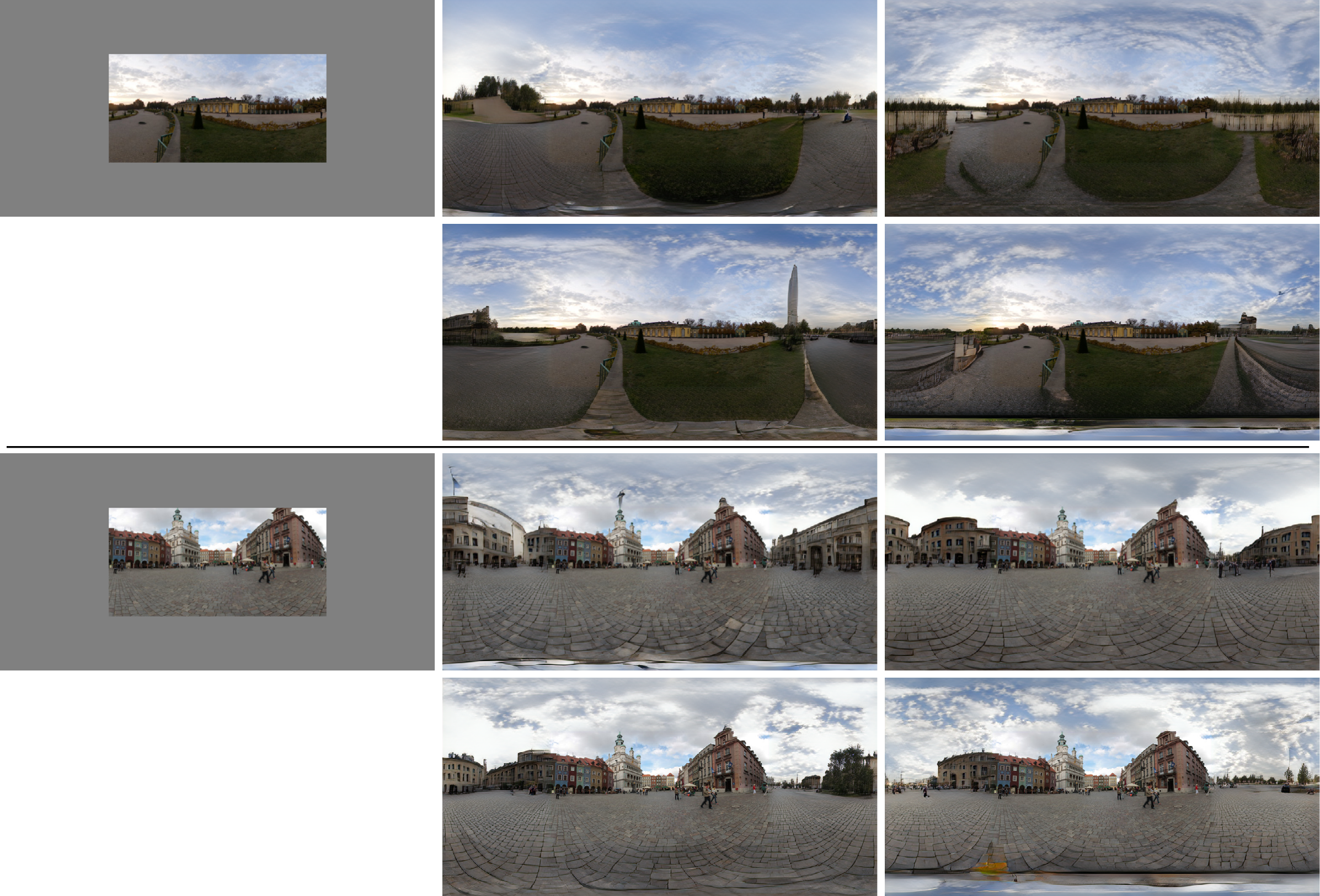}
    \vspace{4mm}
    \caption{Diverse outputs. Our method outputs multiple and diverse results for a single input. Best viewed in color and zoomed in.
    }
    \label{fig:supp_diverse}
\end{figure*}

\begin{figure*}[t]
    \centering
    \includegraphics[width=\linewidth]{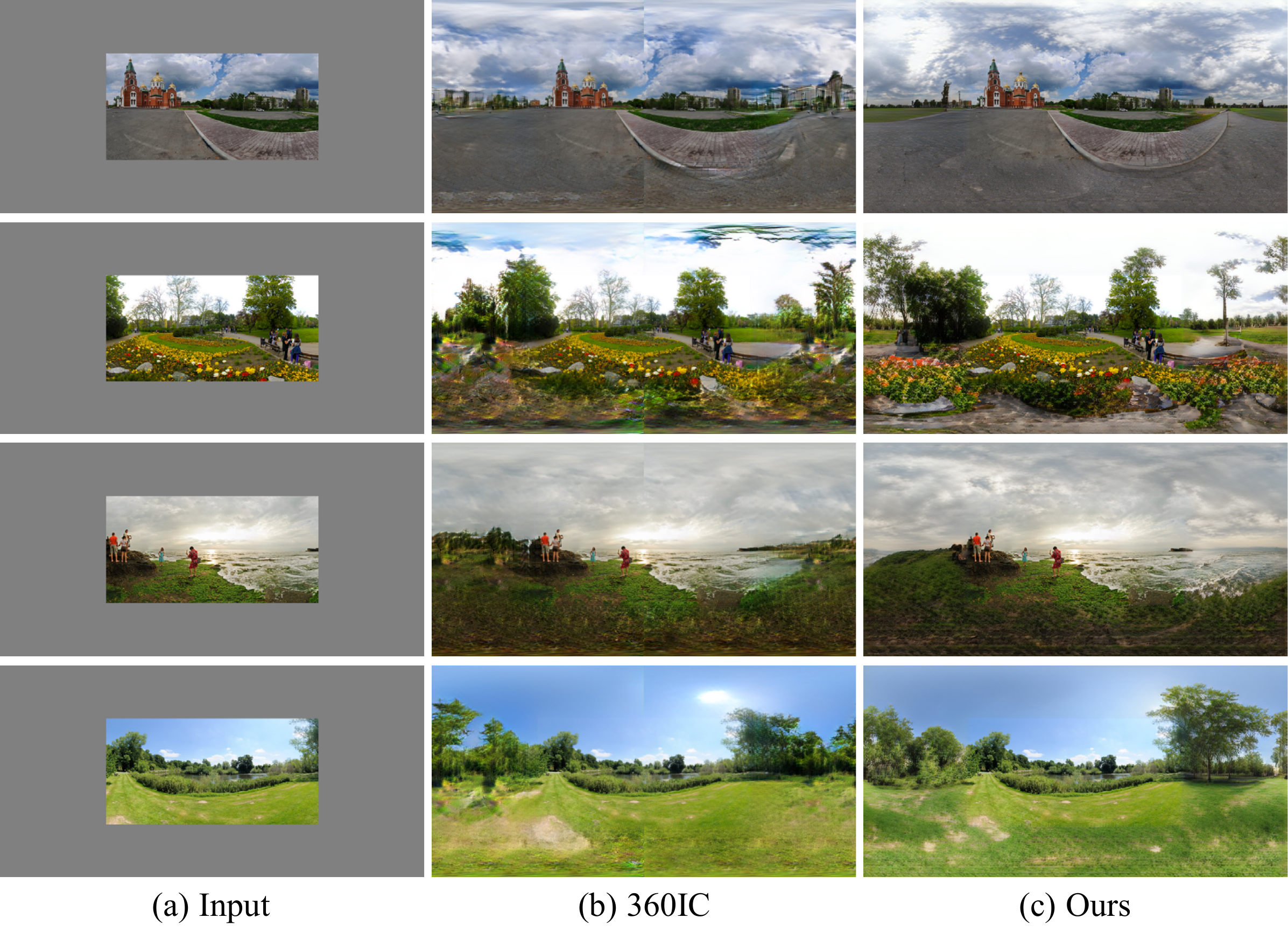}
    \vspace{4mm}
    \caption{Qualitative comparison with 360IC\cite{akimoto2019360}. Best viewed in color and zoomed in.
    }
    \label{fig:supp_vs360ic}
\end{figure*}

\begin{figure*}[t]
    \centering
    \includegraphics[width=\linewidth]{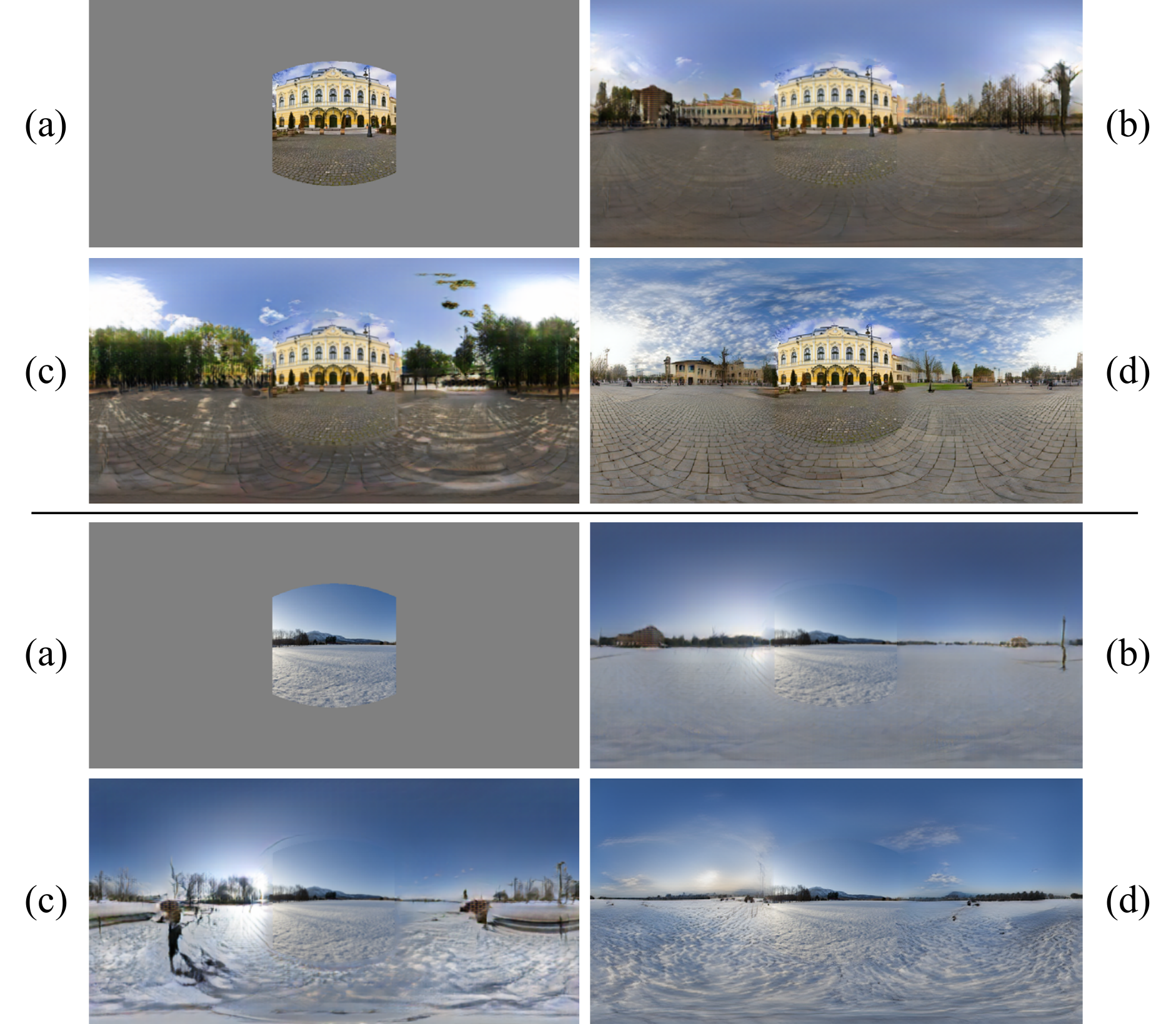}
    \vspace{4mm}
    \caption{Qualitative comparison with SIG-SS \cite{hara2021spherical}. (a) indicates the input image, (b) indicates SIG-SS(rec), (c) indicates SIG-SS(gen), and (d) indicates ours. Best viewed in color and zoomed in.
    }
    \label{fig:supp_vssigss}
\end{figure*}

\begin{figure*}[t]
    \centering
    \includegraphics[width=\linewidth]{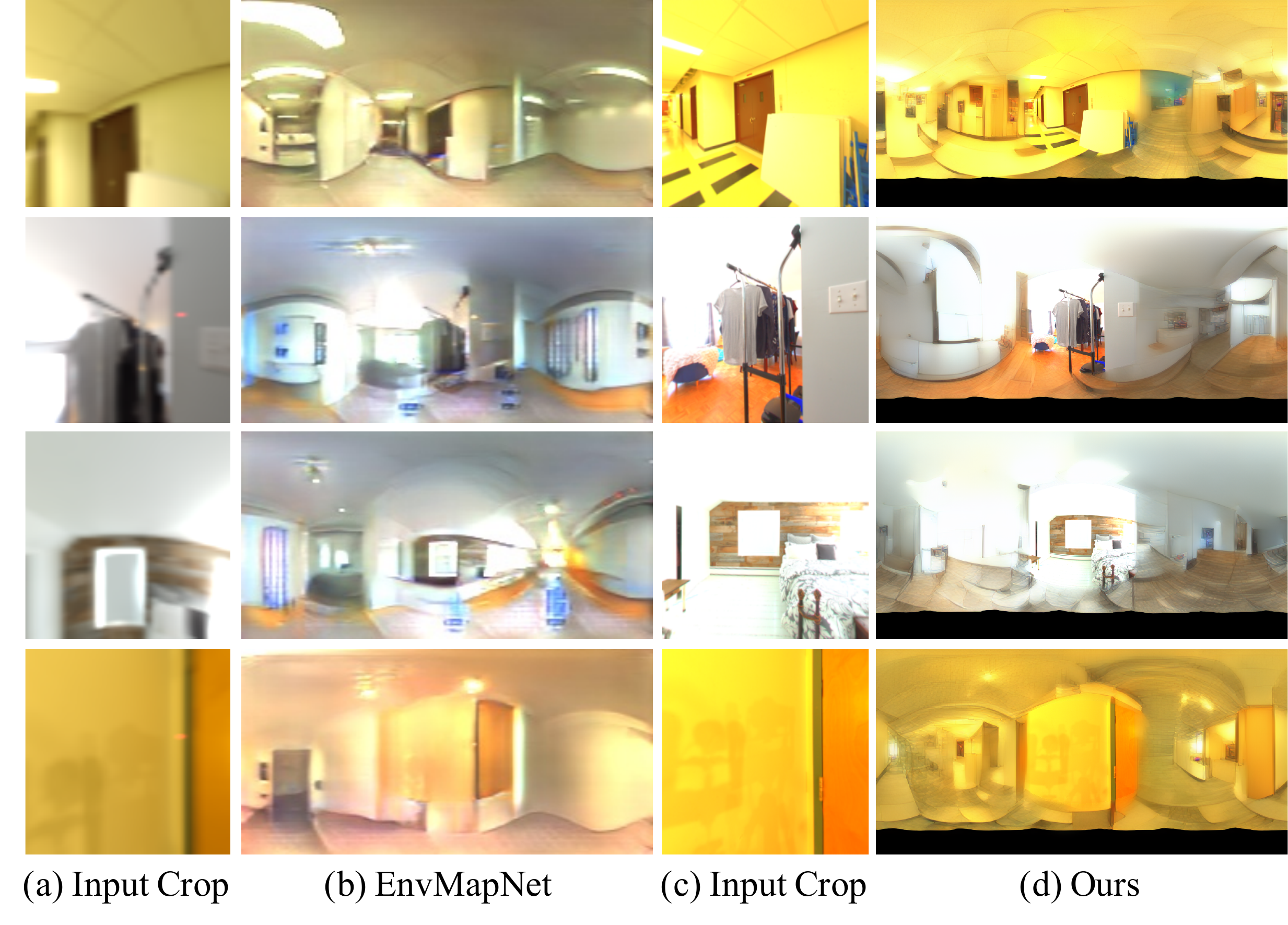}
    \vspace{4mm}
    \caption{Qualitative comparison with EnvMapNet \cite{somanath2021hdr}. We show completion results along with the corresponding input images. Best viewed in color and zoomed in.
    }
    \label{fig:supp_vsenvmapnet}
\end{figure*}

Some of the generated results in Fig.~\ref{fig:limitations_prior} and Fig.~\ref{fig:supp_diverse} have sky pixels generated at positions corresponding to the lower poles. This is due to the training image rather than the proposed method. SUN360 padded the pole with sky pixels or marks to hide the camera equipment that took the 360-degree images. A similar appearance can be seen in Fig.~5 of Akimoto {\it et al.}\cite{akimoto2019360}.

In this paper, we reported equirectangular images. However, because of the inherent distortions of the equirectangular projection, it may be challenging to evaluate generated results intuitively. Therefore, we display equirectangular images as perspective images in a supplementary video. Please refer to the video for reprojection results.

\subsection{Training AdjustmentNet}
\begin{figure*}[t]
\begin{center}
\includegraphics[width=\linewidth]{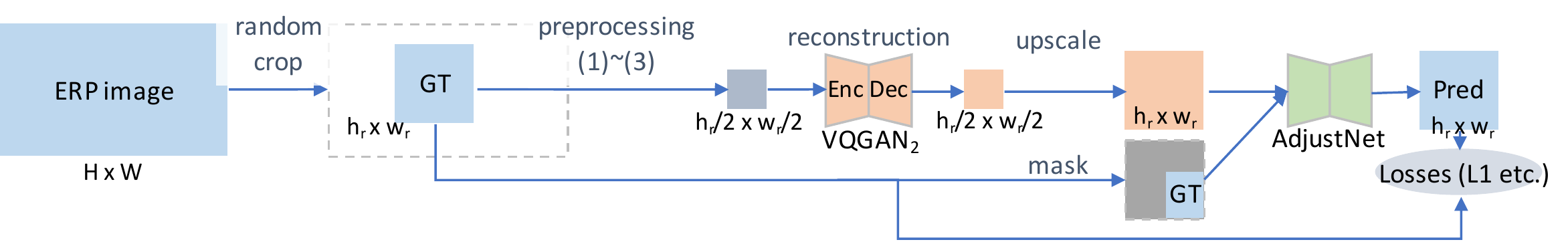}
\end{center}
\caption{Training pipeline for AdjustmentNet. Trained VQGAN$_2$ performs reconstruction, not completion.}
\label{fig:adjust_training}
\end{figure*}

Fig.~\ref{fig:adjust_training} illustrates the training pipeline of AdjustmentNet. See Sec.~\ref{sec:training} for a detailed explanation.

\subsection{Different FOV inputs}
\begin{figure*}[t]
\begin{center}
\includegraphics[width=\linewidth]{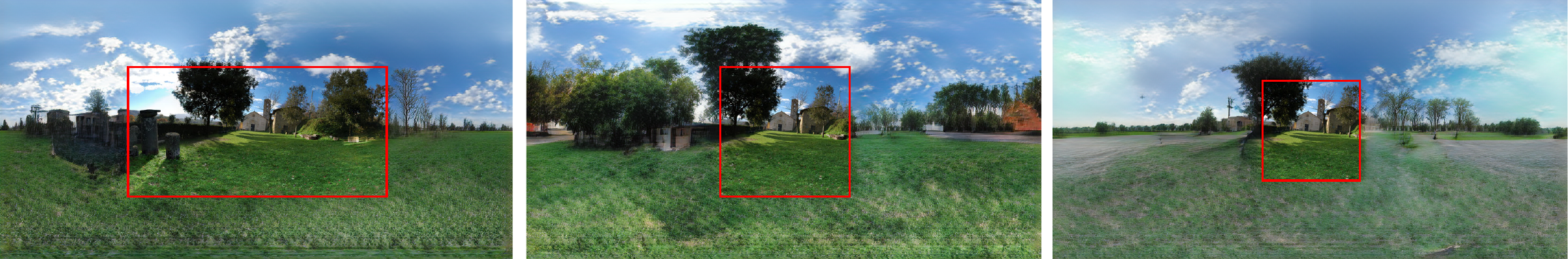}
\end{center}
\caption{Results with different FOV inputs (red boxes). Left to right: $180^{\circ}\times90^{\circ}$ input, $90^{\circ}\times90^{\circ}$ input, and $70^{\circ}\times70^{\circ}$ input.}
\label{fig:different_fov}
\end{figure*}

The ``Ours’’ in Fig.~\ref{fig:comp_360ic} and Fig.~\ref{fig:comp_sigss} are the results of the same trained model but with different FOVs. Furthermore, we show Fig.~\ref{fig:different_fov} to discuss the effect of different FOVs on the results. We perform training at $180^{\circ}\times90^{\circ}$ and inference at multiple FOVs. Fig.~\ref{fig:different_fov} shows that the proposed method is robust to differences in the size of the input region. Furthermore, when the input area is small, the color of the generated texture tends to be lighter. This may be because the input area has less texture and therefore cannot be referenced, resulting in a lack of clarity in the generated texture.

\subsection{Quantity evaluation for diversity}
One of the diversity metrics is the average LPIPS distance, following Wan {\it et al.}\cite{Wan_2021_ICCV}. The score of our proposed method is 0.34 ($180^{\circ}\times90^{\circ}$ input (mask ratio 75\%), 1K inputs, 5 samples per an input, VGG network). We report this score for subsequent works. We cannot evaluate some of the previous methods due to unpublished codes. However, we believe that this quantitative comparison does not change the conclusions of our paper, since 360IC and EnvMapNet are deterministic and SIG-SS is limited only to the type of symmetry.

\section{Supplementary Video}
\label{sec:video}
We publish a video as supplementary material. In this video, we show our diverse completion results and further compare them quantitatively with the results obtained by previous methods through demonstrations. Using the pipeline proposed in Section 5, we demonstrate the camera rotating around virtual objects and also demonstrate a virtual object approaching the camera. Since the original output of EnvMapNet \cite{somanath2021hdr} is an HDR image, but the resulting image is not publicly available, we use the same inverse tone mapping method \cite{liu2020single} as ours to convert the publicly available LDR images into HDR images for the demonstration. We render each video clip by path tracing. We also perform spherical visualization for some of the comparisons. For the visualization, we use {\tt Basic app RICOH THETA}. At the end of the video, we show a demo where we use one completion result to produce the background and ground, and a virtual actor runs on it.

\section*{Acknowledgement}
We thank Teppei Suzuki and Daiki Ito for helpful discussions and comments.

\end{document}